\documentclass[11pt]{article}

\usepackage[final]{acl}

\usepackage{times}
\usepackage{latexsym}
\usepackage[T1]{fontenc}
\usepackage[utf8]{inputenc}
\usepackage{microtype}
\usepackage{inconsolata}
\usepackage{xurl}
\usepackage{graphicx}
\usepackage{xcolor}
\usepackage{booktabs}
\usepackage{fvextra}
\input{prompt-style.tex}
\usepackage{multirow}
\usepackage{amssymb}
\usepackage{xcolor}
\usepackage{tikz}
\usepackage{adjustbox}
\usepackage{tcolorbox}
\usepackage{amsmath} 
\usepackage{placeins}
\usetikzlibrary{trees,arrows.meta}

\definecolor{borderblue}{RGB}{34,51,103}
\definecolor{bggray}{RGB}{245,247,250}

\newcommand{\asc}[1]{\texttt{#1}}

\title{SurfacePrint: Typographic Fingerprints of LLMs}
\author{
  Wentao Zhang$^{1}$ \quad Sajad Ebrahimi$^{2}$ \quad Nima Jamali$^{2}$ \quad Bardia Shirsalimian$^{3}$ \\
  \textbf{Kelly McConvey}$^{2}$ \quad \textbf{Jalehsadat Mahdavimoghaddam}$^{2}$ \quad \textbf{Maksym Taranukhin}$^{4}$ \\
  \textbf{Maura R. Grossman}$^{1}$ \quad \textbf{Vered Shwartz}$^{4}$ \quad \textbf{Ebrahim Bagheri}$^{2}$ \quad \textbf{Yuntian Deng}$^{1}$ \\
  $^{1}$University of Waterloo \quad $^{2}$University of Toronto \\
  $^{3}$Toronto Metropolitan University \quad $^{4}$University of British Columbia \\
  \texttt{\{w564zhan,yuntian\}@uwaterloo.ca}
}

\begin{document}
\maketitle

\begin{abstract}
Detecting AI-generated text from its content becomes harder as language models get better at writing like humans. We show that their outputs still contain distinctive surface-level patterns in punctuation, whitespace, and Markdown formatting. We introduce SurfacePrint, which combines 97 typographic features with a lightweight tree-based classifier. Across 17 LLMs and four domains, it achieves a mean per-generator AUROC of 0.983 without access to the language models and makes predictions in microseconds on a CPU. Cross-model transfer experiments and ablations comparing base and instruction-tuned models show that these formatting patterns are learned during pretraining and strengthened by instruction tuning.
\end{abstract}

\section{Introduction}

Large language models (LLMs) are increasingly trained to match the semantics of human writing. This makes AI-text detection a moving target: the better models become at producing fluent, human-like prose, the less reliable detectors become when they rely on the prose itself. Existing approaches therefore face a difficult tradeoff. Neural classifiers and reference-model methods such as DetectGPT~\citep{mitchell2023detectgpt}, Fast-DetectGPT~\citep{bao2024fastdetectgpt}, Lastde++~\citep{xu2025training}, and DNA-DetectLLM~\citep{zhu2026dna} can be accurate, but many of them require language-model access or expensive scoring, and they target the very signals that newer models are trained to humanize.

We argue that a complementary signal lives one level above the prose. Before a text is read as language, it is also a sequence of surface conventions: punctuation, Markdown, \LaTeX{} delimiters, fenced code, and whitespace. These choices are not merely stylistic. They are shaped by tokenizers, chat templates, instruction tuning, data curation, and rendering defaults. In this sense, surface formatting can act as a production-pipeline fingerprint: not an intrinsic signature of ``AI writing,'' but a residue of how the text was generated.

\begin{figure}[!t]
\centering
\footnotesize
\begin{tikzpicture}[
    every node/.style={align=center, font=\tiny},
    level 1/.style={sibling distance=32mm, level distance=10mm},
    level 2/.style={sibling distance=16mm, level distance=10mm},
    level 3/.style={sibling distance=9mm,  level distance=10mm},
    edge from parent/.style={draw,-{Stealth[length=1.5pt]}},
    decision/.style={draw, rounded corners, inner sep=1.5pt, fill=blue!5},
    leaf/.style={draw, rounded corners, inner sep=1.5pt},
    human/.style={leaf, fill=green!15},
    llm/.style={leaf, fill=red!15},
]
\node [decision] {\asc{**bold**}\\present?}
    child { node [decision] {Unicode\\dash?}
        child { node [decision] {\asc{\textbackslash[\,\textbackslash]}\,display\\math?}
            child { node [human] {\textbf{Human}} edge from parent node[left,font=\tiny] {no} }
            child { node [llm] {\textbf{LLM}} edge from parent node[right,font=\tiny] {yes} }
            edge from parent node[left,font=\tiny] {no}
        }
        child { node [decision] {Inline\\\asc{`\,`\textellipsis`\,`}\,?}
            child { node [llm] {\textbf{LLM}} edge from parent node[left,font=\tiny] {no} }
            child { node [human] {\textbf{Human}} edge from parent node[right,font=\tiny] {yes} }
            edge from parent node[right,font=\tiny] {yes}
        }
        edge from parent node[left,font=\tiny] {no}
    }
    child { node [decision] {Inline\\\asc{`\,`\textellipsis`\,`}\,?}
        child { node [llm] {\textbf{LLM}} edge from parent node[left,font=\tiny] {no} }
        child { node [decision] {Markdown\\escape?}
            child { node [llm] {\textbf{LLM}} edge from parent node[left,font=\tiny] {no} }
            child { node [human] {\textbf{Human}} edge from parent node[right,font=\tiny] {yes} }
            edge from parent node[right,font=\tiny] {yes}
        }
        edge from parent node[right,font=\tiny] {yes}
    };
\end{tikzpicture}
\caption{A depth-3 surface-formatting tree separates human and LLM text with AUROC\,=\,0.7497 using only five typographic and markup tests. The tree illustrates that shallow, interpretable surface conventions already carry model-pipeline signal.}
\label{fig:intro-tree}
\end{figure}
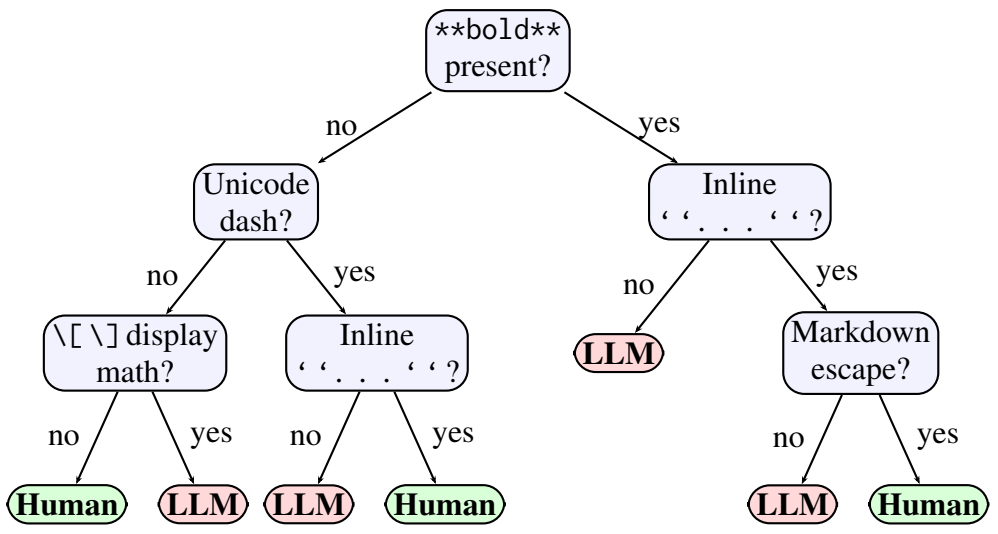

Figure~\ref{fig:intro-tree} makes this observation concrete. A depth-3 decision tree using only five binary glyph tests---whether the text contains \texttt{**bold**}, a Unicode dash, \texttt{\textbackslash[...\textbackslash]} display math, inline code, or a Markdown escape---already separates human from LLM text at AUROC = 0.7497 without any language-model access. The splits are interpretable as pipeline habits. Markdown bold reflects chat-oriented formatting conventions; Unicode dashes reflect model- and provider-specific punctuation preferences; inline code and \LaTeX{} delimiters reflect domain-specific formatting habits learned or amplified during post-training. If a handful of glyph tests already carry this much signal, the full surface layer should carry substantially more.

We operationalize this idea as \textsc{SurfacePrint}, a 97-dimensional feature vector over prose typography, Markdown structure, math and \LaTeX{}, code surface features, and whitespace. Paired with a lightweight tree-based classifier, \textsc{SurfacePrint} provides a fast and interpretable detector that reads the surface dialect of model outputs rather than scoring using another model. Our findings are:
\begin{itemize}
\itemsep1pt
    \item \textbf{Fast detection without language-model access.} A single \textsc{SurfacePrint} classifier achieves 0.983 mean per-generator AUROC across 17 LLMs and four domains, matching or exceeding reference-model baselines at microsecond-level CPU inference.
    \item \textbf{Temporal structure in typographic fingerprints.} Cross-model transfer is asymmetric: detectors trained on older generators transfer forward better than detectors trained on newer generators transfer backward, suggesting that contemporary alignment pipelines have converged on a shared post-2024 surface dialect.
    \item \textbf{Post-training amplifies the surface dialect.} Base models are already typographically separable, but instruction tuning sharply strengthens the fingerprint, with the largest gains appearing in recent post-training pipelines.
\end{itemize}

Together, these results suggest that when LLMs learn to sound human, they may still write in the dialect of their production pipeline. \textsc{SurfacePrint} provides a lightweight way to read that dialect.
Our code is available at \url{https://github.com/yuntian-group/surfaceprint}.

\section{Related Work}
\label{app:related_work}

\paragraph{AI-generated text detection.}
The dominant detection paradigm scores a candidate text under a reference language model. Supervised neural classifiers \citep{solaiman2019release,ippolito2020automatic} train a discriminator on human-vs.-LLM pairs. Zero-shot reference-model methods include DetectGPT \citep{mitchell2023detectgpt}, which scores log-probability curvature under input perturbations; Fast-DetectGPT \citep{bao2024fastdetectgpt}, which replaces expensive perturbations with a conditional probability surrogate; DetectLLM \citep{su-etal-2023-detectllm}, which exploits log-rank statistics; Lastde++ \citep{xu2025training}, which mines token probability sequences; and DNA-DetectLLM \citep{zhu2026dna}, which compares texts to a learned generative ``DNA'' prototype. Watermarking \citep{kirchenbauer2023watermark} provides a complementary, generator-side guarantee. \citet{tang2024science} survey the broader field. Our approach is closest in spirit to GLTR \citep{gehrmann-etal-2019-gltr}, which visualizes per-token rank statistics, but requires no model access at inference and operates entirely on surface formatting.

\paragraph{Stylometry and typographic cues.}
Character- and word-level features have a long history in authorship attribution \citep{uchendu2020authorship,jawahar2020automatic}. Typography itself, in particular the distinction between ASCII and Unicode punctuation, has only recently been observed as an LLM-specific artifact: \citet{freeburg2026fingerprintmarkdowntrainingshapes} note that Markdown training shapes LLM prose toward characteristic em-dash usage. We extend this line of work to a broader 97-dimensional surface representation that covers prose typography, Markdown, math, code, and whitespace, and frame the resulting fingerprint explicitly as a tokenizer- and alignment-pipeline artifact rather than a stylistic choice.

\paragraph{Benchmarks for human--LLM text.}
HC3 \citep{guo-etal-2023-hc3} introduced one of the earliest sizable corpora of paired human and ChatGPT answers and underlies the Reddit-ELI5 split we use. RAID \citep{dugan2024raid} and MAGE \citep{li2024mage} expand coverage across generators, domains, and adversarial conditions; they are complementary to the four datasets of Section~\ref{sec:datasets}, which we source specifically to span technical Q\&A and free-form prose under unified prompts.

\section{SurfacePrint}
\label{sec:intuitions}
\begin{table}[htbp]
    \centering
    \small
    \setlength{\tabcolsep}{4pt}
    \renewcommand{\arraystretch}{1.15}
    \begin{tabular}{@{}l c p{0.52\columnwidth}@{}}
    \toprule
    \textbf{Group} & \textbf{\#} & \textbf{Scope} \\
    \midrule
    Prose typography   & 22           & Unicode punctuation and body-text micro-typography. \\
    Markdown structure & 25           & Markdown delimiter and block-layout conventions. \\
    Math and \LaTeX{}  & 24           & Surface markup of mathematical content. \\
    Code surface       & 20           & Formatting and idiom within source-code blocks. \\
    Whitespace         & \phantom{0}6 & Document-wide invisible-layout signatures. \\
    \midrule
    \textbf{Total}     & \textbf{97}  & \\
    \bottomrule
    \end{tabular}
    \caption{Taxonomy of the 97 \textsc{SurfacePrint} features, grouped by the surface layer they measure. Full feature definitions are provided in Appendix~\ref{app:features}.}
    \label{tab:feature_groups}
    \end{table}

\subsection{Feature taxonomy}
\label{sec:feature_taxonomy}

\textsc{SurfacePrint} represents each document with 97 surface-level features organized into five interpretable groups: prose typography, Markdown structure, math and \LaTeX{}, code surface, and whitespace (Table~\ref{tab:feature_groups}). Each feature is a scalar rate or density in $[0,1]$, computed directly from raw text. Full feature definitions are provided in Appendix~\ref{app:features}.

\subsection{Classifier and variants}
\label{sec:classifier}

We train a random forest on top of the 97-D \textsc{SurfacePrint} vector (Appendix~\ref{app:implementation}). We report two variants that differ only in whitespace handling. \textsc{SP-Raw} strips leading and trailing whitespace before feature extraction. \textsc{SP-WS} removes every ASCII space, tab, carriage return, and line feed before feature extraction.

This distinction separates two kinds of surface signal. Raw whitespace is itself informative: LLM outputs are often more consistently newline-formatted than human writing. However, a classifier trained on raw text can exploit this shortcut and underuse the remaining 91 features. \textsc{SP-WS} neutralizes the strongest whitespace cues, isolating other signal such as punctuation and Markdown.

\subsection{LLM-text detection setup}
\label{sec:detection_setup}

We frame detection as binary human-vs.-LLM classification. Our main detector is a single \textsc{SurfacePrint} classifier trained on 8{,}000 human posts, with 2{,}000 from each dataset in Section~\ref{sec:datasets}, and 470 outputs from each of the 17 generators in Section~\ref{sec:models}, for $470 \times 17 = 7{,}990$ LLM examples. We then evaluate this same classifier on the held-out test split of every generator--dataset cell and report per-generator AUROC in Table~\ref{tab:baseline_statistical_llama}.

This protocol follows the per-generator reporting convention used by FastDetectGPT~\citep{bao2024fastdetectgpt}, Lastde++~\citep{xu2025training}, and DNA-DetectLLM~\citep{zhu2026dna}, making the results directly comparable to the reference-model baselines in Section~\ref{sec:baselines}. The small number of training examples per generator is intentional: it measures how much per-model labeling is needed to extend the detector to a newly released generator.

\begin{table*}[htbp]
    \centering
    \footnotesize
    \setlength{\tabcolsep}{4pt}
    \renewcommand{\arraystretch}{1.05}
    \begin{tabular}{l|ccccccc|cc}
    \toprule
    Generator Model & FDGPT & LogLik & LogRank & Entropy & LRR & Lastde++ & DNA-LLM & SP-WS & SP-Raw \\
    \midrule
    \texttt{opt-iml-1.3b} & 0.966 & 0.894 & 0.896 & 0.840 & 0.855 & 0.959 & \textbf{0.984} & 0.725 & 0.902 \\
    \texttt{gpt-3.5-turbo} & 0.953 & 0.786 & 0.790 & 0.752 & 0.785 & 0.924 & \textbf{0.970} & 0.727 & 0.937 \\
    \texttt{llama-2-7b} & \textbf{0.997} & 0.872 & 0.886 & 0.848 & 0.932 & 0.994 & 0.995 & 0.811 & 0.979 \\
    \texttt{claude-3-haiku} & \textbf{0.989} & 0.850 & 0.859 & 0.831 & 0.887 & 0.981 & 0.984 & 0.812 & 0.986 \\
    \texttt{gpt-4o-mini} & 0.976 & 0.801 & 0.808 & 0.784 & 0.827 & 0.964 & 0.969 & 0.928 & \textbf{0.996} \\
    \texttt{llama-3.1-8b} & 0.997 & 0.927 & 0.934 & 0.878 & 0.936 & 0.995 & \textbf{0.998} & 0.852 & 0.992 \\
    \texttt{claude-3.5-haiku} & 0.928 & 0.697 & 0.702 & 0.694 & 0.706 & 0.912 & 0.937 & 0.821 & \textbf{0.992} \\
    \texttt{gemini-2.0-flash} & 0.966 & 0.783 & 0.787 & 0.768 & 0.785 & 0.956 & 0.976 & 0.884 & \textbf{0.990} \\
    \texttt{gpt-4.1-mini} & 0.951 & 0.747 & 0.753 & 0.740 & 0.754 & 0.936 & 0.954 & 0.933 & \textbf{0.994} \\
    \texttt{claude-sonnet-4} & 0.964 & 0.754 & 0.762 & 0.758 & 0.783 & 0.954 & 0.953 & 0.950 & \textbf{0.996} \\
    \texttt{gemini-2.5-flash} & 0.955 & 0.774 & 0.779 & 0.762 & 0.790 & 0.949 & 0.965 & 0.904 & \textbf{0.988} \\
    \texttt{olmo-3-7b} & 0.901 & 0.679 & 0.690 & 0.668 & 0.735 & 0.911 & 0.889 & 0.940 & \textbf{0.987} \\
    \texttt{qwen3.5-9b} & 0.954 & 0.736 & 0.746 & 0.729 & 0.783 & 0.952 & 0.928 & 0.969 & \textbf{0.997} \\
    \texttt{claude-sonnet-4-6} & 0.908 & 0.691 & 0.702 & 0.699 & 0.730 & 0.893 & 0.870 & 0.968 & \textbf{0.997} \\
    \texttt{gpt-5.4-mini} & 0.894 & 0.692 & 0.698 & 0.674 & 0.706 & 0.893 & 0.870 & 0.973 & \textbf{0.997} \\
    \texttt{gemma-4-e4b} & 0.888 & 0.642 & 0.644 & 0.648 & 0.656 & 0.893 & 0.847 & 0.958 & \textbf{0.990} \\
    \texttt{gemini-3.5-flash} & 0.956 & 0.734 & 0.740 & 0.724 & 0.757 & 0.942 & 0.922 & 0.967 & \textbf{0.998} \\
    \midrule
    \textbf{Average} & 0.950 & 0.768 & 0.775 & 0.753 & 0.789 & 0.942 & 0.942 & 0.889 & \textbf{0.983} \\
    \bottomrule
    \end{tabular}
    \caption{Detector AUROC by generator model (FastDetectGPT, Lastde++, and DNA-DetectLLM also use Llama-3.1-8B as reference and observer). Methods: \texttt{FDGPT}=FastDetectGPT, \texttt{LogLik}=Log-likelihood, \texttt{LogRank}=LogRank, \texttt{Entropy}=Entropy, \texttt{LRR}=LRR, \texttt{Lastde++}=Lastde++, \texttt{DNA-LLM}=DNA-DetectLLM, \texttt{SP-WS}=SurfacePrint (whitespace norm.), \texttt{SP-Raw}=SurfacePrint (raw norm.).}
    \label{tab:baseline_statistical_llama}
\end{table*}

\section{Experimental Setup}
\label{sec:setup}

\paragraph{Datasets}
\label{sec:datasets}
We evaluate on four datasets spanning two broad regimes where surface formatting is especially salient: technical question answering and free-form prose. \textbf{StackExchange-Programming} is newly sourced from the HuggingFace StackExchange preference data~\citep{h4stackexchange}, restricted to programming-related topics and to posts before November 2022, predating the public release of ChatGPT. \textbf{StackExchange-Math} is taken from \citet{zhang2024stackmathqa} and covers mathematics Q\&A with substantial \LaTeX{} content. \textbf{Reddit-ELI5} comes from HC3~\citep{guo-etal-2023-hc3}, and \textbf{Reddit-WritingPrompt} from \citet{fan-etal-2018-hierarchical}. Both are dominated by long-form prose. Each dataset contributes 2{,}000 prompts for training, 500 for validation, and 1{,}000 for testing. %

\paragraph{Generator Models}
\label{sec:models}
We evaluate 17 LLMs released between December 2022 and May 2026, covering both proprietary and open-weight systems from OpenAI, Anthropic, Google, Meta, and other open-weight labs. The set includes models across five successive generations of frontier and open-weight systems; the full list, with release dates, is provided in Appendix~Table~\ref{tab:model_names}. For each dataset in Section~\ref{sec:datasets}, we generate completions from every model using the prompts in Appendix~\ref{app:prompts}.

\paragraph{Detection Baselines}
\label{sec:baselines}

We compare \textsc{SurfacePrint} against seven statistical and reference-model detectors: Entropy~\citep{gehrmann2019gltr}, Likelihood~\citep{hashimoto-etal-2019-unifying}, LogRank~\citep{jawahar2020automatic}, LRR~\citep{su-etal-2023-detectllm}, Fast-DetectGPT, Lastde++, and DNA-DetectLLM. All baselines use \texttt{Llama-3.1-8B-Instruct}~\citep{grattafiori2024llama3herdmodels} as the scoring model. Fast-DetectGPT, Lastde++, and DNA-DetectLLM additionally use \texttt{Llama-3.1-8B} as the reference or observer model. Hyperparameters and implementation details for both the baselines and our random forest classifier are reported in Appendix~\ref{app:implementation}.

\section{Results and Discussion}
\label{sec:results}

\subsection{Main Results}
\label{sec:main_results}

Table~\ref{tab:baseline_statistical_llama} shows that \textsc{SP-Raw} reaches a mean per-generator AUROC of \textbf{0.983}, taking the top score on 12 of 17 target generators. It outperforms every reference-model baseline while requiring no language-model access at inference: FastDetectGPT reaches 0.950, Lastde++ and DNA-DetectLLM reach 0.942, and the simpler statistical baselines fall between 0.75 and 0.79. Since \textsc{SurfacePrint} uses only a 97-dimensional feature vector and a lightweight tree-based classifier, inference runs at microsecond-level CPU cost.

The trend across model generations is especially striking. Reference-model detectors are strongest on earlier generators and degrade on the 2024--2026 frontier: DNA-DetectLLM falls to 0.870 on \texttt{gpt-5.4-mini} and \texttt{claude-sonnet-4.6}, and to 0.847 on \texttt{gemma-4-e4b}. In contrast, \textsc{SP-Raw} exceeds 0.99 AUROC on every evaluated flagship released since \texttt{gpt-4o-mini}, peaking at 0.998 on \texttt{gemini-3.5-flash}. Thus, in our setting, surface separability increases across model generations even as reference-model separability declines.

Raw whitespace is a strong surface cue, so we also evaluate \textsc{SP-WS}, which normalizes the most direct whitespace shortcuts. Even after this normalization, \textsc{SP-WS} retains a mean AUROC of 0.889, outperforming all four statistical baselines and remaining within 0.06 AUROC of FastDetectGPT. This shows that the signal is not only newline formatting: punctuation, Markdown, math, and code-formatting conventions alone carry substantial information about contemporary LLM outputs.

\subsection{Temporal Effects}
\label{sec:cross_model}

\begin{figure}[h]
    \centering
    \includegraphics[width=1.0\linewidth]{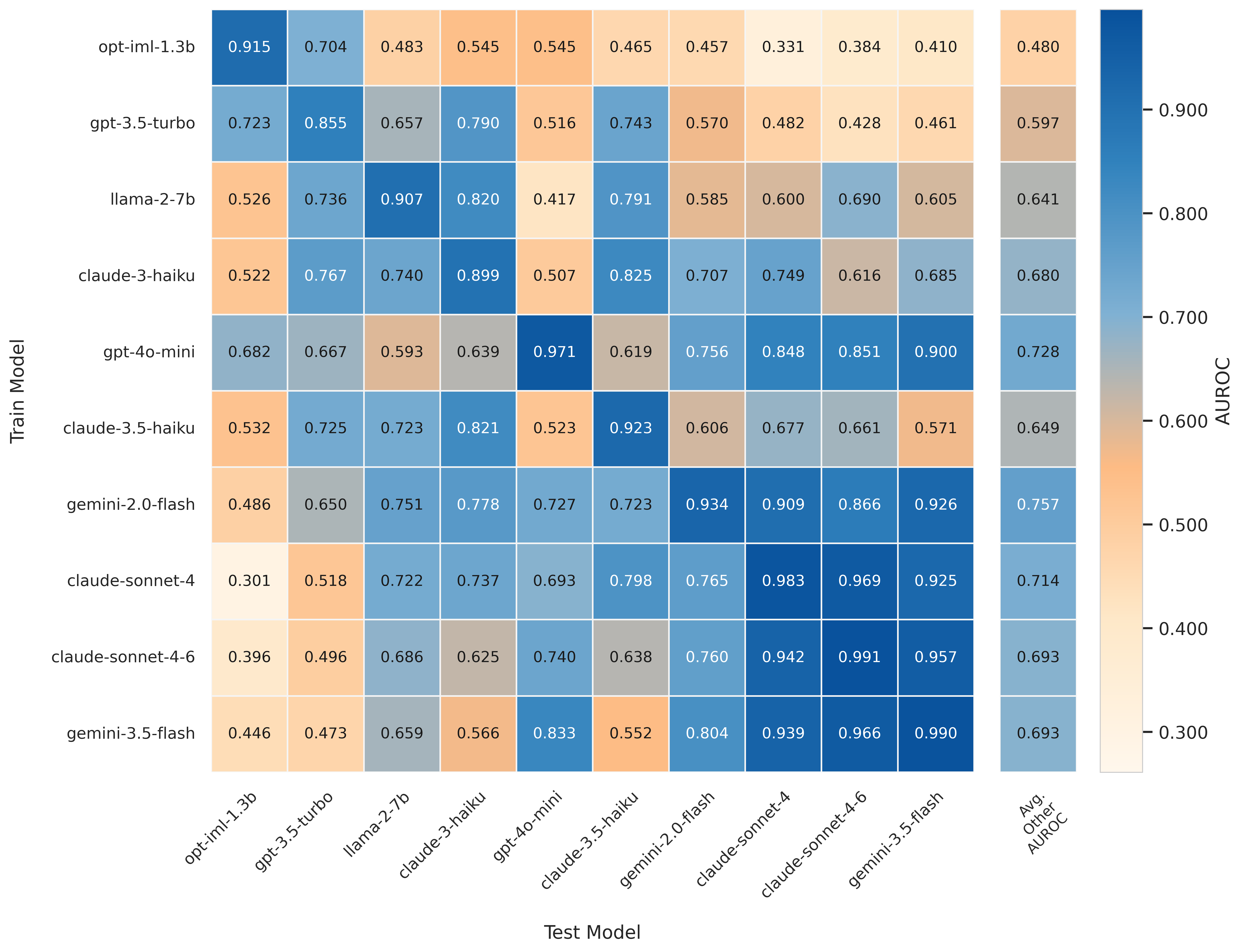}
    \caption{Cross-model generalization of \textsc{SurfacePrint} under whitespace normalization. Rows are training generators and columns are test generators, ordered by release date; the rightmost strip reports mean off-diagonal AUROC. Transfer is temporally asymmetric: older-model fingerprints transfer forward better than newer-model fingerprints transfer backward.}
    \label{fig:one_to_all_sampled}
\end{figure}

To test whether typographic fingerprints transfer across generators, we train a separate \textsc{SP-WS} classifier for each generator $g$, using 2{,}000 human posts and 2{,}000 outputs from $g$, and evaluate it on outputs from every other generator. Appendix~\ref{app:one_to_all_full} reports the full $17 \times 17$ transfer matrix.

Figure~\ref{fig:one_to_all_sampled} reveals a temporally asymmetric pattern. Transfer degrades sharply backward in time but only mildly forward. For example, a detector trained on \texttt{gpt-4o-mini} retains AUROC between 0.848 and 0.972 on every later flagship, but drops to 0.593-0.682 on pre-2024 generators. Similarly, a detector trained on \texttt{claude-sonnet-4} reaches 0.969 on \texttt{claude-sonnet-4-6}, but only 0.301 on \texttt{opt-iml-1.3b}. The row-mean off-diagonal AUROC, shown in the rightmost strip, rises with release date, peaks at 0.757 for \texttt{gemini-2.0-flash}, and then plateaus near 0.70 across 2025-2026 flagships. This suggests that recent alignment pipelines have converged toward a shared surface dialect across providers.

\begin{table}[htbp]
    \centering

    \footnotesize
    \setlength{\tabcolsep}{4pt}
    \renewcommand{\arraystretch}{1.08}
    \begin{tabular}{lccc}
    \toprule
    \textbf{Model} & \textbf{Base} & \textbf{Instruct} & \textbf{$\Delta$} \\
    \midrule
    \texttt{llama-2-7b} & 0.810 & 0.804 & \textcolor{red!75!black}{-0.006} \\
    \texttt{llama-3.1-8b} & 0.744 & 0.880 & \textcolor{teal!70!black}{+0.136} \\
    \texttt{olmo-3-7b} & 0.709 & 0.958 & \textcolor{teal!70!black}{+0.249} \\
    \texttt{qwen3.5-9b} & 0.840 & 0.978 & \textcolor{teal!70!black}{+0.138} \\
    \texttt{gemma-4-e4b} & 0.754 & 0.963 & \textcolor{teal!70!black}{+0.209} \\
    \midrule
    \textbf{Average} & 0.771 & 0.917 & \textcolor{teal!70!black}{+0.145} \\
    \bottomrule
    \end{tabular}
    \caption{\textsc{SurfacePrint} detection AUROC for base and instruction-tuned checkpoints from the same model families. Instruction tuning generally amplifies the typographic fingerprint. $\Delta=\text{instruct}-\text{base}$.}
    \label{tab:base_vs_instruct}
    \end{table}

\subsection{Post-Training Effects}
\label{sec:base_vs_instruct}

To locate where the typographic signature emerges, we compare \textsc{SP-WS} detection AUROC on five open-weight model families for which both base and instruction-tuned checkpoints are available (Table~\ref{tab:base_vs_instruct}). Base models are already separable, with a mean AUROC of 0.771, indicating that part of the fingerprint is seeded during pretraining. Instruction tuning then amplifies the signal, raising mean AUROC to 0.917 ($\Delta = +0.145$).

The largest gains occur in recent model families: $+0.249$ for \texttt{olmo-3-7b}, $+0.209$ for \texttt{gemma-4-e4b}, and $+0.138$ for \texttt{qwen3.5-9b}. The only exception is \texttt{llama-2-7b} ($\Delta = -0.006$), whose 2023-era chat template adds little surface structure beyond the base model. Together with the temporal plateau in Section~\ref{sec:cross_model}, these results suggest that the modern fingerprint is shaped less by pretraining corpora alone than by contemporary post-training and chat-formatting pipelines.

\section{Conclusion}

We introduced \textsc{SurfacePrint}, a 97-dimensional fingerprint of surface-level features such as punctuation, whitespace, Markdown, \LaTeX{}, and code formatting. Across 17 LLMs and four domains, a lightweight tree-based classifier using these features achieves a mean per-generator AUROC of 0.983 without language-model access and at microsecond-level CPU cost. Beyond detection accuracy, our results show that these formatting patterns are model-specific and are strengthened by posttraining: cross-model transfer degrades sharply backward in time but only mildly forward, and base-instruct ablations show that instruction tuning amplifies these patterns. Our results suggest that surface-level features can support fast and interpretable detection of AI-generated text. We hope that \textsc{SurfacePrint} can serve as a simple baseline for future work on AI-text detection.

\section*{Limitations}

\paragraph{Supervision overhead.}
Unlike training-free detectors such as Fast-DetectGPT, Lastde++, and DNA-DetectLLM, \textsc{SurfacePrint} requires a modest labeled training set (8{,}000 human posts and 470 examples per generator as described in Section~\ref{sec:detection_setup}). Inference remains at microsecond CPU latency, but the method cannot be deployed zero-shot on newly released generators without fresh labels.

\paragraph{Temporal decay.}
The fingerprint evolves with alignment pipelines. Detectors trained on older generators lose power on newer ones (Section~\ref{sec:cross_model}), and the modern signature is largely a post-training artifact (Section~\ref{sec:base_vs_instruct}). As laboratories diversify or sanitize surface conventions, periodic re-training will be necessary.

\paragraph{Adversarial robustness.}
Surface formatting is trivially editable: a motivated adversary can largely defeat \textsc{SurfacePrint} with a simple normalization pass over whitespace, Unicode punctuation, Markdown, and math delimiters. We make no claim of robustness against such attacks, the detector targets non-adversarial detection at scale, where its speed and interpretability matter most.

\section*{Ethical Considerations}

AI text detection carries risks of false accusations.
Our detector is a lightweight heuristic, not a definitive tool.
Most humans do not produce the distinguishing feature, and false positives are expected.
It should never be sole evidence in high-stakes decisions.

\section*{Acknowledgments}
This work was supported by the Safeguarding Courts from Synthetic AI Content Solution Network, funded through the Canadian AI Safety Institute (CAISI) Research Program at CIFAR. Yuntian Deng acknowledges support from the Natural Sciences and Engineering Research Council of Canada (NSERC) through Discovery Grant RGPIN-2024-05178. We also thank the Vector Institute for its support.

\appendix

\FloatBarrier
\section{Generator Models}
\label{app:models}

\begin{table}[htbp]
    \centering
    \small
    \begin{tabular}{@{}p{0.68\columnwidth}|l@{}}
    \toprule
    Model Name & Release Date \\
    \hline
    \texttt{opt-iml-1.3b}~\citep{iyer2023optimlscalinglanguagemodel} & 2022-12-22 \\
    \texttt{gpt-3.5-turbo}~\citep{openai2023chatgpt} & 2023-03-01 \\
    \texttt{llama-2-7b-chat}~\citep{touvron2023llama2openfoundation} & 2023-07-18 \\
    \texttt{claude-3-haiku}~\citep{anthropic2024claude3} & 2024-03-13 \\
    \texttt{gpt-4o-mini}~\citep{openai2024gpt4omini} & 2024-07-18 \\
    \texttt{llama-3.1-8b-instruct}~\citep{grattafiori2024llama3herdmodels} & 2024-07-23 \\
    \texttt{claude-3.5-haiku}~\citep{anthropic2024claude35haiku} & 2024-11-04 \\
    \texttt{gemini-2.0-flash}~\citep{google2024gemini2} & 2025-02-05 \\
    \texttt{gpt-4.1-mini}~\citep{openai2025gpt41} & 2025-04-14 \\
    \texttt{claude-sonnet-4}~\citep{anthropic2025claude4} & 2025-05-22 \\
    \texttt{gemini-2.5-flash}~\citep{google2025gemini25} & 2025-06-17 \\
    \texttt{olmo-3-7b-instruct}~\citep{olmo2026olmo3} & 2025-11-21 \\
    \texttt{qwen3.5-9b}~\citep{yang2025qwen3technicalreport} & 2026-02-16 \\
    \texttt{claude-sonnet-4.6} & 2026-02-17 \\
    \texttt{gpt-5.4-mini} & 2026-03-17 \\
    \texttt{gemma-4-e4b-it}~\citep{google2026gemma4} & 2026-04-02 \\
    \texttt{gemini-3.5-flash} & 2026-05-19 \\
    \bottomrule
    \end{tabular}
    \caption{Models Tested}
    \label{tab:model_names}
\end{table}

\FloatBarrier
\section{SurfacePrint Feature List}
\label{app:features}
\begin{table*}[htbp]
\centering

\small
\begin{tabular}{@{}p{0.35\linewidth}p{\dimexpr0.65\linewidth-2\tabcolsep\relax}@{}}
\toprule
\textbf{Feature} & \textbf{Description} \\
\midrule
\texttt{apostrophe\_curly\_rate} &
Share of curly apostrophe (U+2019) among all apostrophe characters. \\
\texttt{single\_quote\_pair\_unicode\_rate} &
Share of Unicode single-quote pairs (U+2018--U+2019) among single-quoted spans. \\
\texttt{double\_quote\_pair\_unicode\_rate} &
Share of Unicode double-quote pairs (U+201C--U+201D) among double-quoted spans. \\
\texttt{dash\_emdash\_rate} &
Share of em dash (U+2014) among em dash, en dash, and hyphen forms. \\
\texttt{dash\_endash\_rate} &
Share of en dash (U+2013) among em dash, en dash, and hyphen forms. \\
\texttt{ellipsis\_unicode\_rate} &
Share of Unicode ellipsis (U+2026) among ellipsis positions (including \texttt{...}). \\
\texttt{minus\_unicode\_rate} &
Within inline-math spans, share of Unicode minus (U+2212) over ASCII hyphen / Unicode minus. \\
\texttt{nbsp\_rate\_per\_word} &
Count of non-breaking, narrow-no-break, and thin spaces per word. \\
\texttt{invisible\_char\_presence} &
Binary indicator: any zero-width character, BOM, or word joiner present. \\
\texttt{french\_spacing\_rate} &
Share of \texttt{; : ! ?} preceded by a space among all such punctuation marks. \\
\texttt{comma\_fullwidth\_rate} &
Share of fullwidth (CJK) comma among all comma characters. \\
\texttt{bullet\_star\_rate} &
Share of \texttt{*} among line-leading bullet markers. \\
\texttt{quote\_unicode\_rate} &
Share of curly quote codepoints among all straight and curly quote marks. \\
\texttt{dash\_unicode\_rate} &
Share of Unicode dash forms among Unicode dashes plus ASCII hyphens. \\
\texttt{special\_space\_rate\_per\_word} &
Count of NBSP, thin/hair/zero-width, and word-joiner spaces per word. \\
\texttt{post\_vowel\_apostrophe\_curly\_rate} &
Curly-apostrophe share after vowels (rough contraction context). \\
\texttt{post\_consonant\_apostrophe\_curly\_rate} &
Curly-apostrophe share after consonants. \\
\texttt{trailing\_space\_rate\_per\_line} &
Share of lines ending with literal space or tab. \\
\texttt{double\_space\_after\_period\_rate} &
Share of period-following spaces that use two or more spaces. \\
\texttt{blank\_line\_between\_paragraphs\_rate} &
Blank-line separators per non-empty prose line. \\
\texttt{emoji\_presence} &
Binary indicator: any emoji or pictographic decoration codepoint present. \\
\texttt{emoji\_rate\_per\_word} &
Count of emoji / pictographic decoration codepoints per word. \\
\bottomrule
\end{tabular}
\caption{Group~A: Prose typography features (22). Each feature is a scalar in $[0,1]$ unless noted; ratio features use a 0.5 default when the denominator is zero.}
\label{tab:features-a}
\end{table*}

\begin{table*}[htbp]
\centering
\small
\begin{tabular}{@{}p{0.35\linewidth}p{\dimexpr0.65\linewidth-2\tabcolsep\relax}@{}}
\toprule
\textbf{Feature} & \textbf{Description} \\
\midrule
\texttt{bold\_double\_star\_rate} &
Share of \texttt{**bold**} delimiters among bold markers (\texttt{**} vs.\ \texttt{\_\_}). \\
\texttt{italic\_star\_rate} &
Share of \texttt{*italic*} among italic delimiters after masking bold spans. \\
\texttt{heading\_atx\_rate} &
Share of ATX headings (\texttt{\# H1}) among all Markdown headings. \\
\texttt{code\_fence\_lang\_tag\_rate} &
Share of opening code fences that carry a language tag. \\
\texttt{code\_fence\_backtick\_rate} &
Share of backtick fences (\texttt{\textasciigrave\textasciigrave\textasciigrave}) over tilde fences (\texttt{\textasciitilde\textasciitilde\textasciitilde}). \\
\texttt{link\_inline\_rate} &
Share of inline \texttt{[text](url)} links among all Markdown link forms. \\
\texttt{numbered\_dot\_rate} &
Share of \texttt{1.} form among numbered-list openers. \\
\texttt{hardbreak\_double\_space\_rate} &
Share of double-space hard breaks (\texttt{\textasciitilde\textasciitilde\textbackslash n}) vs.\ backslash-newline vs.\ blank line. \\
\texttt{hardbreak\_backslash\_rate} &
Share of hard breaks written as backslash-newline. \\
\texttt{hardbreak\_blank\_line\_rate} &
Share of paragraph breaks represented by blank lines. \\
\texttt{bullet\_dash\_rate} &
Share of \texttt{-} among line-leading bullet markers. \\
\texttt{bullet\_plus\_rate} &
Share of \texttt{+} among line-leading bullet markers. \\
\texttt{bullet\_unicode\_rate} &
Share of Unicode bullets among Markdown-like bullet markers. \\
\texttt{numbered\_roman\_rate} &
Share of roman-numeral numbered-list openers. \\
\texttt{numbered\_alpha\_rate} &
Share of alphabetic-letter numbered-list openers. \\
\texttt{blank\_line\_before\_heading\_rate} &
Share of Markdown headings preceded by a blank line or start of file. \\
\texttt{code\_fence\_tilde\_rate} &
Share of tilde fences (\texttt{\textasciitilde\textasciitilde\textasciitilde}) over all fenced code blocks. \\
\texttt{indented\_code\_block\_rate} &
Share of four-space/tab indented code blocks vs.\ fenced blocks. \\
\texttt{inline\_code\_double\_backtick\_rate} &
Share of inline code spans delimited with double backticks. \\
\texttt{image\_inline\_rate} &
Share of images using inline \texttt{![alt](url)} syntax. \\
\texttt{blockquote\_nested\_rate} &
Share of blockquote lines with nesting depth greater than one. \\
\texttt{blank\_line\_after\_blockquote\_rate} &
Share of blockquote blocks followed by a blank line or EOF. \\
\texttt{table\_cell\_padding\_rate} &
Share of Markdown table cells padded with spaces on both sides. \\
\texttt{table\_center\_align\_rate} &
Share of Markdown table alignment cells using centered \texttt{:---:} form. \\
\texttt{markdown\_escape\_density\_per\_kb} &
Density of Markdown escapes for \texttt{*}, \texttt{\_}, and backticks (per KB). \\
\bottomrule
\end{tabular}
\caption{Group~B: Markdown structure features (25).}
\label{tab:features-b}
\end{table*}

\begin{table*}[htbp]
\centering
\small
\begin{tabular}{@{}p{0.35\linewidth}p{\dimexpr0.65\linewidth-2\tabcolsep\relax}@{}}
\toprule
\textbf{Feature} & \textbf{Description} \\
\midrule
\texttt{tex\_quote\_convention\_rate} &
Share of TeX-convention \texttt{\textasciigrave\textasciigrave...\textasciigrave\textasciigrave} / \texttt{\textasciigrave...\textasciigrave} pairs among quoted spans. \\
\texttt{emdash\_ligature\_rate} &
Share of \texttt{---} TeX ligature among em-dash positions. \\
\texttt{endash\_ligature\_rate} &
Share of \texttt{--} TeX ligature among en-dash positions. \\
\texttt{inline\_math\_dollar\_rate} &
Share of \texttt{\$...\$} among inline-math delimiters (\texttt{\$} vs.\ \texttt{\textbackslash(...\textbackslash)}). \\
\texttt{display\_math\_dollar\_rate} &
Share of \texttt{\$\$...\$\$} among display-math delimiters. \\
\texttt{greek\_command\_rate} &
Share of \LaTeX{} Greek-letter commands vs.\ literal Unicode Greek characters. \\
\texttt{math\_op\_command\_rate} &
Share of \LaTeX{} math-operator commands vs.\ literal Unicode equivalents. \\
\texttt{subscript\_brace\_rate} &
Share of braced subscripts \texttt{x\_\{...\}} vs.\ bare \texttt{x\_i}. \\
\texttt{superscript\_brace\_rate} &
Share of braced superscripts \texttt{x\^{}\{...\}} vs.\ bare \texttt{x\^{}2}. \\
\texttt{fraction\_command\_rate} &
Share of fractions written as \texttt{\textbackslash frac\{a\}\{b\}}. \\
\texttt{fraction\_parenthesized\_slash\_rate} &
Share of slash fractions written as \texttt{(a)/(b)}. \\
\texttt{bracket\_left\_right\_rate} &
Share of bracket delimiters using \texttt{\textbackslash left}/\texttt{\textbackslash right} sizing. \\
\texttt{emphasis\_textit\_rate} &
Share of \LaTeX{} emphasis written as \texttt{\textbackslash textit\{\}} rather than \texttt{\textbackslash emph\{\}}. \\
\texttt{cite\_plain\_rate} &
Share of citation commands using plain \texttt{\textbackslash cite\{\}}. \\
\texttt{cite\_textual\_rate} &
Share of natbib citations using textual \texttt{\textbackslash citet\{\}} form. \\
\texttt{display\_math\_bracket\_rate} &
Share of display math delimited with \texttt{\textbackslash[...\textbackslash]}. \\
\texttt{display\_math\_environment\_rate} &
Share of display math using \texttt{equation}/\texttt{align}-like environments. \\
\texttt{operator\_spacing\_density\_per\_equation} &
Density of TeX spacing commands per inline/display equation marker. \\
\texttt{boxed\_expression\_rate} &
Density of \texttt{\textbackslash boxed\{\}} uses per equation marker. \\
\texttt{therefore\_implies\_rate} &
Density of \texttt{\textbackslash therefore}/\texttt{\textbackslash implies} proof connectors per equation. \\
\texttt{equation\_alignment\_ampersand\_rate} &
Share of display environments that contain alignment ampersands. \\
\texttt{equation\_linebreak\_density} &
Density of \LaTeX{} equation line breaks (\texttt{\textbackslash\textbackslash}) per display environment. \\
\texttt{reference\_autoref\_rate} &
Share of references using \texttt{\textbackslash autoref\{\}}. \\
\texttt{reference\_eqref\_rate} &
Share of references using \texttt{\textbackslash eqref\{\}}. \\
\bottomrule
\end{tabular}
\caption{Group~C: Math and \LaTeX{} surface-choice features (24). LaTeX-conditioned features return 0.5 when no \LaTeX{} markers are detected.}
\label{tab:features-c}
\end{table*}

\begin{table*}[htbp]
\centering
\small
\begin{tabular}{@{}p{0.35\linewidth}p{\dimexpr0.65\linewidth-2\tabcolsep\relax}@{}}
\toprule
\textbf{Feature} & \textbf{Description} \\
\midrule
\texttt{code\_string\_single\_quote\_rate} &
Share of code string literals using single quotes. \\
\texttt{code\_indent\_2space\_rate} &
Share of indented code lines using a two-space indent. \\
\texttt{code\_indent\_4space\_rate} &
Share of indented code lines using a four-space indent. \\
\texttt{code\_indent\_tab\_rate} &
Share of indented code lines using tab indentation. \\
\texttt{code\_comment\_hash\_rate} &
Share of comments using Python/shell \texttt{\#} form. \\
\texttt{code\_comment\_slash\_rate} &
Share of comments using \texttt{//} form. \\
\texttt{code\_comment\_block\_rate} &
Share of comments using \texttt{/* ... */} form. \\
\texttt{code\_comment\_sql\_dash\_rate} &
Share of comments using SQL-style \texttt{--} form. \\
\texttt{type\_annotation\_spaced\_rate} &
Share of type annotations written as \texttt{x: int} rather than \texttt{x:int}. \\
\texttt{fstring\_double\_quote\_rate} &
Share of formatting expressions using double-quoted f-strings. \\
\texttt{fstring\_single\_quote\_rate} &
Share of formatting expressions using single-quoted f-strings. \\
\texttt{format\_method\_rate} &
Share of string formatting using \texttt{.format()}. \\
\texttt{percent\_format\_rate} &
Share of string formatting using \texttt{\%} interpolation. \\
\texttt{code\_trailing\_comma\_rate} &
Share of collection/call literals containing a trailing comma. \\
\texttt{docstring\_presence\_rate} &
Docstring count per Python \texttt{def}/\texttt{class} block in code. \\
\texttt{docstring\_triple\_double\_rate} &
Share of docstrings using triple-double quotes. \\
\texttt{docstring\_first\_line\_summary\_rate} &
Share of docstrings whose first content line is a short summary sentence. \\
\texttt{lambda\_over\_def\_rate} &
Share of callable definitions written as \texttt{lambda} rather than \texttt{def}. \\
\texttt{listcomp\_over\_loop\_rate} &
Share of list-building opportunities using list comprehensions. \\
\texttt{import\_blankline\_group\_rate} &
Share of adjacent import blocks separated by blank lines. \\
\bottomrule
\end{tabular}
\caption{Group~D: Code-surface features (20). Computed over fenced and indented code blocks extracted from the document.}
\label{tab:features-d}
\end{table*}

\begin{table*}[htbp]
\centering
\small
\begin{tabular}{@{}p{0.35\linewidth}p{\dimexpr0.65\linewidth-2\tabcolsep\relax}@{}}
\toprule
\textbf{Feature} & \textbf{Description} \\
\midrule
\texttt{crlf\_newline\_rate} &
Share of newlines encoded as CRLF rather than LF. \\
\texttt{tokenizer\_fragmentation\_proxy\_per\_kb} &
Light CPT proxy: non-ASCII and markup-punctuation fragments per KB. \\
\texttt{leading\_tab\_rate} &
Share of indented lines whose leading indentation contains tabs. \\
\texttt{indent\_consistency\_rate} &
Dominant indentation-style share among tabs, two spaces, and four spaces. \\
\texttt{final\_newline\_present} &
Binary indicator: text ends with a newline. \\
\texttt{section\_blank\_line\_rate} &
Share of section/list boundaries separated by a blank line. \\
\bottomrule
\end{tabular}
\caption{Group~E: Whitespace and invisible-character features (6).}
\label{tab:features-e}
\end{table*}

\clearpage
\onecolumn
\section{Prompts}
\label{app:prompts}
\begin{figure}[!h]
    \centering
   \begin{tcolorbox}[colframe=borderblue, colback=bggray, left=1mm, right=1mm, top=0.5mm, bottom=0.5mm]
    {\fvset{fontsize=\scriptsize, breaklines, breakanywhere, breaksymbolleft={}}\input{prompts/wp_prompt_base.tex}}
    \end{tcolorbox}
    \caption{Prompt template of base model for WritingPrompt dataset}
\end{figure}

\begin{figure}[!h]
    \centering
   \begin{tcolorbox}[colframe=borderblue, colback=bggray, left=1mm, right=1mm, top=0.5mm, bottom=0.5mm]
    {\fvset{fontsize=\scriptsize, breaklines, breakanywhere, breaksymbolleft={}}\input{prompts/reddit_eli5_base.tex}}
    \end{tcolorbox}
    \caption{Prompt template of base model for Reddit ELI5}
\end{figure}

\begin{figure}[!h]
    \centering
   \begin{tcolorbox}[colframe=borderblue, colback=bggray, left=1mm, right=1mm, top=0.5mm, bottom=0.5mm]
    {\fvset{fontsize=\scriptsize, breaklines, breakanywhere, breaksymbolleft={}}\input{prompts/stackexchange_math_base.tex}}
    \end{tcolorbox}
    \caption{Prompt template of base model for StackExchange Math}
\end{figure}

\begin{figure}[!h]
    \centering
   \begin{tcolorbox}[colframe=borderblue, colback=bggray, left=1mm, right=1mm, top=0.5mm, bottom=0.5mm]
    {\fvset{fontsize=\scriptsize, breaklines, breakanywhere, breaksymbolleft={}}\input{prompts/stackexchange_programming_base.tex}}
    \end{tcolorbox}
    \caption{Prompt template of base model for StackExchange Programming}
\end{figure}

\begin{figure}[!h]
    \centering
   \begin{tcolorbox}[colframe=borderblue, colback=bggray, left=1mm, right=1mm, top=0.5mm, bottom=0.5mm]
    {\fvset{fontsize=\scriptsize, breaklines, breakanywhere, breaksymbolleft={}}\input{prompts/wp_prompt.tex}}
    \end{tcolorbox}
    \caption{Prompt template of instructional fine-tuned model for WritingPrompt dataset}
\end{figure}

\begin{figure}[!h]
    \centering
   \begin{tcolorbox}[colframe=borderblue, colback=bggray, left=1mm, right=1mm, top=0.5mm, bottom=0.5mm]
    {\fvset{fontsize=\scriptsize, breaklines, breakanywhere, breaksymbolleft={}}\input{prompts/reddit_eli5.tex}}
    \end{tcolorbox}
    \caption{Prompt template of instructional fine-tuned model for Reddit ELI5}
\end{figure}

\begin{figure}[!h]
    \centering
   \begin{tcolorbox}[colframe=borderblue, colback=bggray, left=1mm, right=1mm, top=0.5mm, bottom=0.5mm]
    {\fvset{fontsize=\scriptsize, breaklines, breakanywhere, breaksymbolleft={}}\input{prompts/stackexchange_math.tex}}
    \end{tcolorbox}
    \caption{Prompt template of instructional fine-tuned model for StackExchange Math}
\end{figure}

\begin{figure}[!h]
    \centering
   \begin{tcolorbox}[colframe=borderblue, colback=bggray, left=1mm, right=1mm, top=0.5mm, bottom=0.5mm]
    {\fvset{fontsize=\scriptsize, breaklines, breakanywhere, breaksymbolleft={}}\input{prompts/stackexchange_programming.tex}}
    \end{tcolorbox}
    \caption{Prompt template of instructional fine-tuned model for StackExchange Programming}
\end{figure}

\clearpage
\section{Full One-to-All Cross-Model Generalization Results}
\label{app:one_to_all_full}

\begin{figure}[h]
    \centering
    \includegraphics[width=1.0\linewidth]{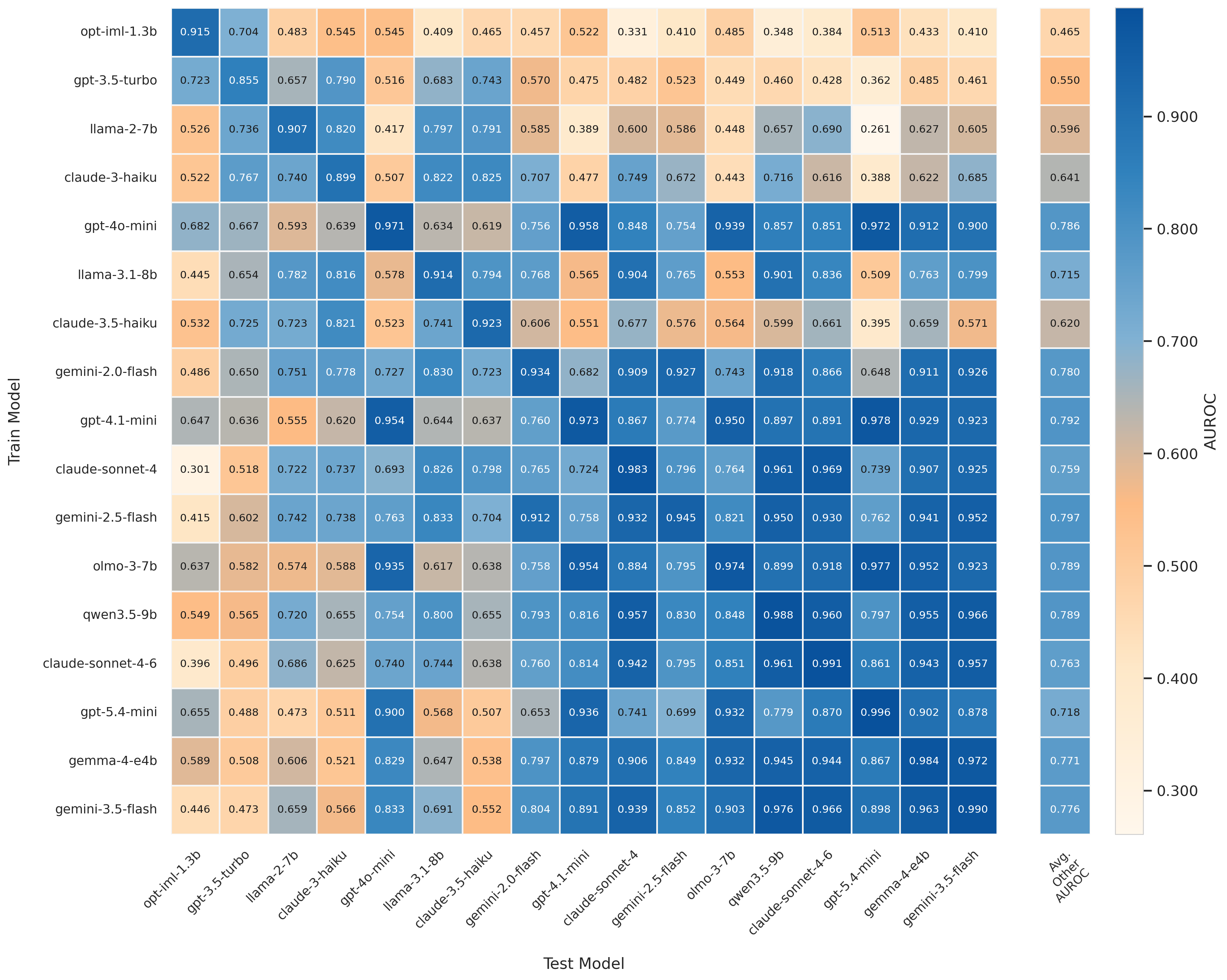}
    \caption{Full one-to-all cross-model generalization (AUROC) under SP-WS over all 17 generators. Rows are train models and columns are test models, both ordered by release date from earliest (top/left) to most recent (bottom/right).}
    \label{fig:one_to_all_full}
\end{figure}

\twocolumn

\clearpage
\section{Implementation Details}
\label{app:implementation}

\paragraph{Data generation.}
We generate completions from open-weight models locally with vLLM, using one or two NVIDIA H100 GPUs depending on availability. For closed-source models, we query the official Anthropic, OpenAI, and Google APIs directly, and use OpenRouter as a unified gateway for the remaining proprietary endpoints.

\paragraph{Model sizes and compute budget.}
The open-weight generators were OPT (1.3B parameters), Llama 2 and OLMo 3 (7B each), Llama 3.1 (8B), Gemma 4 E4B (8B including embeddings, 4.5B effective), and Qwen3.5 (9B). The released dataset contains 167{,}020 responses generated locally and 153{,}956 collected through APIs. We estimate that local generation and the seven reference-model baselines together required approximately 100 H100 GPU-hours. SurfacePrint feature extraction, training, and inference ran on CPUs.

\paragraph{Feature extraction.}
The 97-dimensional SurfacePrint vector is computed entirely on CPU with pure-Python regular expressions and Unicode lookups, with no language-model calls or tokenizer invocations. Ratio-style features fall back to a $0.5$ default when their denominator is zero (e.g., \LaTeX{}-conditioned features on a document containing no math).

\paragraph{Random Forest classifier.}
Both \texttt{SP-Raw} and \texttt{SP-WS} are trained with \texttt{sklearn.ensemble.RandomForest\-Classifier}. We use an ensemble of \textbf{300 trees}, each grown to a maximum depth of \textbf{6} with a minimum of \textbf{3 samples per leaf}; this shallow, regularized configuration is enough to fit the 97-dimensional surface vector while keeping inference at microsecond CPU latency. To absorb the residual class imbalance between the 8{,}000 human posts and the $470 \times 17$ machine examples, we set \texttt{class\_weight = "balanced"}, which reweights each class inversely to its frequency. All experiments use a fixed \texttt{random\_state = 0} for reproducibility, and tree fitting is parallelized across all available CPU cores (\texttt{n\_jobs = -1}). All other arguments take their scikit-learn defaults (Gini impurity, $\sqrt{d}$ features per split, bootstrap sampling with replacement). The same hyperparameters are reused without modification across every experiment in the paper, including the per-generator detection in Section~\ref{sec:main_results}, the one-to-all transfer matrix of Section~\ref{sec:cross_model}, and the base-versus-instruct comparison of Section~\ref{sec:base_vs_instruct}.

\paragraph{Reference-model baselines.}
Entropy, Likelihood, LogRank, LRR, Fast-DetectGPT, Lastde++, and DNA-DetectLLM are run with the official implementations released by their authors and their recommended default hyperparameters. All seven use \texttt{Llama-3.1-8B-Instruct}~\citep{grattafiori2024llama3herdmodels} as the scoring model; Fast-DetectGPT, Lastde++, and DNA-DetectLLM additionally use the base \texttt{Llama-3.1-8B} as the reference/observer model. Scoring is run on a single NVIDIA H100 GPU with \texttt{float16} weights, and per-generator AUROC is computed on the same held-out test split used by SurfacePrint to keep all numbers in Table~\ref{tab:baseline_statistical_llama} directly comparable.

\section{Use of AI Assistants}
\label{app:ai_assistance}

We used AI tools in two limited ways during preparation of this work. Claude Code was used to polish the language of the manuscript. All scientific claims, experimental results, and technical content were written and verified by the authors, and the agent was used only for stylistic and grammatical editing. For implementation, we used Cursor, an AI-assisted coding environment, to help write and refactor experiment and analysis code, and all code was reviewed, tested, and validated by the authors.

\end{document}